\documentclass[conference]{IEEEtran}
\IEEEoverridecommandlockouts
\usepackage{graphicx} 
\usepackage{pifont}
\usepackage{amsmath}
\usepackage{amssymb}  
\usepackage{amsfonts}
\usepackage{algorithmic}
\usepackage{array}
\usepackage{multicol}

\usepackage{textcomp}
\usepackage{stfloats}
\usepackage{url}
\usepackage{verbatim}
\usepackage{graphicx}
\usepackage{caption}
\usepackage{subcaption}
\usepackage[ruled, vlined, linesnumbered]{algorithm2e}
\usepackage{cite}
\usepackage{booktabs}
\usepackage{amssymb,amsfonts,bm}
\usepackage{amsmath,tikz}
\usepackage{color}
\usepackage{mathtools}
\usepackage[hidelinks]{hyperref} 
\usepackage{rotating}
\usepackage{blkarray}
\usepackage{physics}
\usetikzlibrary{arrows}
\usepackage{makecell}

\usepackage{amsthm}
\usepackage{comment}
\usepackage{multirow}
\usepackage{geometry}
\graphicspath{{images/}}
\usepackage[font=small]{caption}
\begin{document}


  
\title{\LARGE \bf
STARC: See-Through-Wall Augmented Reality Framework \\ for Human–Robot Collaboration in Emergency Response

\author{Shenghai Yuan$^{*}$,~\IEEEmembership{Member,~IEEE}, 
Weixiang Guo$^{*}$, 
Tianxin Hu, 
Yu Yang, \\
Jinyu Chen, 
Rui Qian, 
Zhongyuan Liu, 
and Lihua Xie,~\IEEEmembership{Fellow,~IEEE}%
\thanks{$^{*}$ Equal Contribution.}%
\thanks{All authors are with the School of Electrical and Electronic Engineering, Nanyang Technological University, 50 Nanyang Avenue, Singapore 639798.}%
\thanks{Emails: \{shyuan, elhxie\}@ntu.edu.sg}%
}





}

\maketitle

\begin{abstract}
In emergency response missions, first responders must navigate cluttered indoor environments where occlusions block direct line-of-sight, concealing both life-threatening hazards and victims in need of rescue. We present STARC, a see-through AR framework for human–robot collaboration that fuses mobile-robot mapping with responder-mounted LiDAR sensing. A ground robot running LiDAR–inertial odometry performs large-area exploration and 3D human detection, while helmet- or handheld-mounted LiDAR on the responder is registered to the robot’s global map via relative pose estimation. This cross-LiDAR alignment enables consistent first-person projection of detected humans and their point clouds—rendered in AR with low latency—into the responder’s view. By providing real-time visualization of hidden occupants and hazards, STARC enhances situational awareness and reduces operator risk. Experiments in simulation, lab setups, and tactical field trials confirm robust pose alignment, reliable detections, and stable overlays, underscoring the potential of our system for fire-fighting, disaster relief, and other safety-critical operations. Code and design will be open-sourced upon acceptance.
\end{abstract}


\section{Introduction}
\begin{figure}[t]
  \centering
  \includegraphics[width=\linewidth]{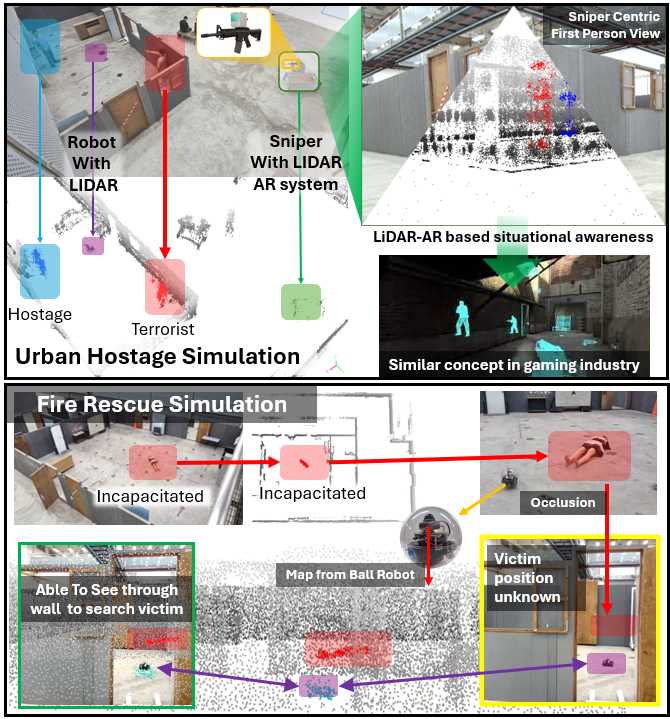}
  \caption{Example use cases of LiDAR–AR situational awareness. Top (hostage scene): a mobile robot with LiDAR builds a global map, while a sniper equipped with a LiDAR–AR system receives through-wall projections of terrorists and hostages, similar in spirit to visualization concepts in the gaming industry. Bottom (fire rescue scene): a ball robot maps a cluttered indoor environment, enabling rescuers to see through occlusions and locate incapacitated victims otherwise hidden behind walls.}
  \label{fig:motivation}
\end{figure}

In recent years, there has been growing interest in technologies that enhance indoor situational awareness for emergency response \cite{schroth2024emergency}. Scenarios such as fires, building collapses, or hostage environments share a common challenge: responders operate in cluttered, occlusion-heavy spaces with limited line-of-sight. Hidden threats endanger their safety while victims remain difficult to locate. Uncertain layouts, obstructed occupants, and unseen hazards make reliable situational understanding urgent, underscoring the need for robotic systems that can scout, map, and interpret complex environments before responders are placed at risk (See Fig.\ref{fig:motivation}).


\textbf{Existing work} has primarily focused on teleoperated robots \cite{erat2018drone}, which provide only 2D video feeds without geometric context, require continuous operator supervision, and lack the ability to convey spatially grounded information. Other approaches \cite{karanam20173d,schroth2024emergency} have investigated wireless or radar-based through-wall sensing, but these methods often lack robustness and generalization across different environments and materials, limiting their applicability in diverse real-world scenarios. More recent systems have explored AR/VR interfaces \cite{chen20243d} to support higher-level human–robot interaction, yet they are unable to provide real-time, occlusion-aware localization of hidden threats or people in need of rescue. Advances in embodied AI, such as vision–language or multimodal agents \cite{sun2025frontiernet}, show promise for richer semantic scene understanding; however, their reliance on heavy models and cloud-scale inference makes them unsuitable for time-critical missions, where latency and resource constraints are critical. To date, there remains no practical, lightweight framework that can deliver real-time 3D situational awareness in occlusion-rich environments under the operational demands of fire-fighting, disaster response, or hostage environments.

The \textbf{challenges} are threefold. First, responders need perception that can reveal people or hazards hidden by walls or obstacles. Second, this information must be presented in a way that aligns with the responder’s own view, for example, through AR overlays. Third, the system must operate under real-time constraints, ensuring that perception, mapping, and visualization remain responsive in safety-critical missions.

To address these challenges, we develop an AR-driven human–robot cooperation framework for situational awareness. A mobile robot with LiDAR–inertial odometry (LIO) builds a global map and detects humans in 3D, while operator-mounted LiDAR provides local geometry that is aligned to the map via relative pose estimation. Instead of re-implementing low-level modules, we integrate proven components—FAST-LIO2 for odometry, ICP/NDT for registration, and PointPillars for detection—to focus on validating the overarching concept: real-time projection of 3D human detections into the operator’s AR view. This cross-platform registration and visualization pipeline enables occlusion-tolerant mapping, robustness to reflective surfaces, and intuitive AR overlays, thereby reducing risks to law enforcement personnel and enhancing situational awareness in safety-critical indoor missions. To our knowledge, no prior work combines operator-mounted LiDAR, robot global mapping, and real-time AR visualization of human detections through cross-LiDAR registration in a ROS~2 system.

\begin{itemize}
\item \textbf{Operator–robot AR framework:} We present \emph{STARC}, the first system to fuse robot mapping with operator-mounted LiDAR sensing, enabling see-through AR overlays for situational awareness in safety-critical indoor missions.

\item \textbf{Task-driven cross-view integration:} Rather than treating cross-LiDAR alignment as an end in itself, we repurpose it as a means to anchor operator viewpoints into the robot map, allowing human detections to be projected consistently into first-person AR views.

\item \textbf{Validation across mission scenarios:} We establish a multi-stage evaluation platform—simulation, controlled lab trials, and tactical field tests—with quantitative metrics, demonstrating the feasibility of AR-enhanced human–robot collaboration for previously unaddressed safety-critical tasks.




\end{itemize}
Note: \noindent\textit{All experiments use safe prop devices under non-hazardous conditions, and the framework is intended solely for situational awareness in search, rescue, and other safety-critical operations.}

\section{Related Works}
\subsection{Robotics for Safety-Critical Reconnaissance}

Recent research has advanced along three directions: air–ground collaboration, teleoperation, and AR/VR-mediated HRI.
UAV–UGV frameworks achieve accurate mapping in GNSS-denied settings~\cite{yue2022aerial}, but focus only on geometry without projecting semantics to operators.
ColAG~\cite{li2024colag} enables cooperative mobility by assigning waypoints, yet depends on reliable communication and offers no AR threat cues.
Teleoperation with haptic feedback~\cite{coffey2022collaborative} aids collision avoidance but remains geometry-bound and lacks occlusion-aware semantics.
AR-based HRI~\cite{qiu2020shared} supports shared holographic workspaces, but validation is limited to tabletop tasks without LiDAR-based reasoning.

Across these strands, systems still (1) emphasize geometry over human/threat semantics, (2) lack alignment of robot maps with operator-mounted sensors, and (3) rarely provide fused 3D cues in real time. These gaps motivate our unified AR framework for occlusion-robust situational awareness.

\subsection{LiDAR-Inertial Odometry and Cross-View Registration}
Recent LIO research has progressed along two fronts: faster odometry and more robust registration. On the odometry side, FAST-LIO2~\cite{xu2022fast}, IG-LIO~\cite{chen2024ig}, and HCTO~\cite{li2024hcto} deliver high-rate updates with reliable filtering, while methods such as UA-MPC~\cite{li2025ua} and graph-based LiDAR bundle adjustment~\cite{li2025graph} push robustness further. These works produce accurate single-platform trajectories but are not designed for scalable multi-agent operations. More recent Swarm-LIO \cite{zhu2023swarmlio,zhu2023swarmlio2} systems extend perception to multi-robot settings, yet they are typically validated only with cooperative targets marked by IR-reflective tape for identification, limiting general applicability.

For registration, ICP and NDT~\cite{biber2003normal, magnusson2009evaluation} remain practical on edge devices due to their efficiency and robustness. Recent methods such as STD \cite{yuan2023STD}, Outram \cite{yin2024outram}, Segregator \cite{Yin2023segregator}, and TripletLoc introduce semantic cues or deep descriptors, while global-representation approaches like Scan Context \cite{kim2018scan}, BEVPlace \cite{luo2023bevplace}, and BEVLIO \cite{cai2025bevlio} achieve long-range matching through robust representations. However, these techniques usually assume wide-FOV spinning LiDARs, structured outdoor environments, or clear building facades; they are often unstable indoors, and unsuitable for directional sensors such as Livox Avia. In cluttered safety-critical sites, such assumptions are rarely satisfied, limiting their practical deployment.

\subsection{Human Detection and Human-Robot Interaction}

Human detection for mobile robots has traditionally been dominated by vision-based approaches, since RGB images provide rich appearance and texture cues for identifying people and their actions. However, in safety-critical missions such as firefighting or search-and-rescue, lighting conditions are often poor or dynamic, making pure camera-based pipelines unreliable. Recent works have therefore shifted toward LiDAR-based person detection, which is robust to illumination but suffers from sparsity, low angular resolution, and the lack of color cues. These limitations hinder fine-grained human recognition. State-of-the-art detectors in autonomous driving benchmarks demonstrate progress on LiDAR pedestrian detection \cite{Shi_2020_CVPR,Yin_2021_CVPR}, and fusion-based pipelines further improve stability by leveraging complementary RGB features \cite{Lang_2019_CVPR,Qin_2022_ECCV}. Nevertheless, these approaches are primarily optimized for automotive scenes rather than close-range robotic reconnaissance, and their heavy backbones challenge real-time, on-edge deployment in field robots.

In parallel, research on human–robot interaction (HRI) has explored AR/VR-mediated interfaces. 
Recent works demonstrate that virtual reality teleoperation can reduce operator workload and 
enable customizable shared control for complex manipulation tasks \cite{Luo_2024_IROS}, while 
mixed-reality supervision and telepresence interfaces improve situational awareness in outdoor 
field robotics \cite{walker2021mixed}. Beyond ground robots, drone-augmented vision has also been 
proposed to extend an operator’s field of view into occluded or hidden areas \cite{erat2018drone}. 
These systems highlight the benefits of immersive interfaces, but they largely remain focused on 
geometric or viewpoint augmentation. They do not yet provide synchronized 3D human semantics or 
fuse operator-mounted sensing with a robot’s global map, and their latency/compute demands still 
pose challenges for deployment in safety-critical edge settings.

In summary, existing perception and HRI pipelines seldom provide a ROS~2-native, real-time solution that (i) leverages LiDAR robustly for human detection under degraded visibility, (ii) mitigates sparsity and semantic limitations to support action-level understanding, and (iii) streams aligned AR overlays to operators with actionable latency in safety-critical indoor missions.

\begin{figure*}[t]
  \centering
  \includegraphics[width=\linewidth]{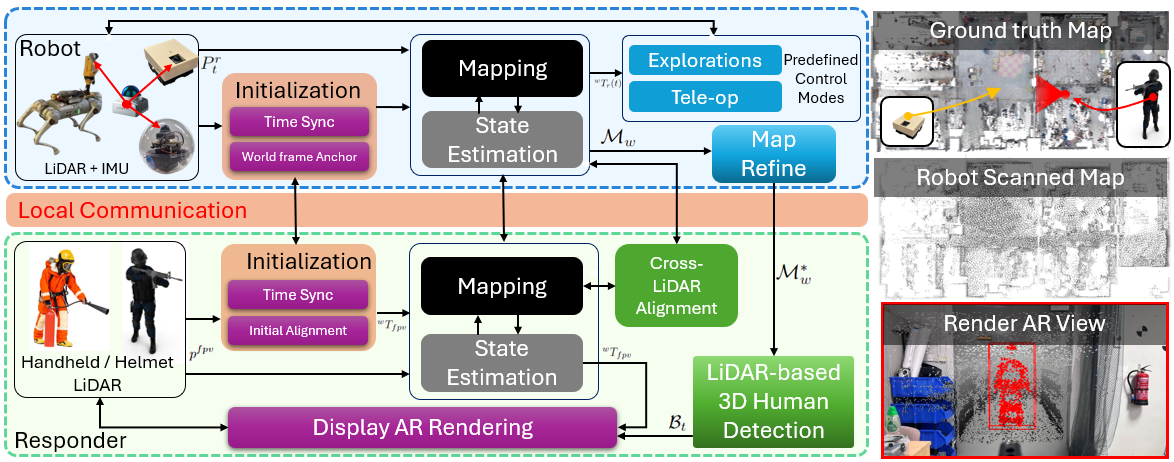}
    \vspace{-10pt}
  \caption{Overview of the STARC system. The robot performs state estimation, mapping, and exploration/teleoperation to build a global map, while the responder carries a handheld/helmet LiDAR for local state estimation. Cross-LiDAR alignment enables the fusion of both perspectives, allowing detected humans to be projected into the responder’s AR view for real-time situational awareness.}
  \label{fig:flowchart}  \vspace{-10pt}
\end{figure*}

\section{Proposed Framework}
\subsection{Problem Definition}

We consider a safety-critical indoor environment represented by a 3D world frame $\mathcal{W} \subset \mathbb{R}^3$. A mobile robot $\mathcal{R}$ is deployed ahead of human operators to construct a global map $\mathcal{M}_w$ using LiDAR--Inertial Odometry (LIO). A first responder $\mathcal{H}_p$ carries a helmet- or handhold-mounted LiDAR, whose role is to perceive the environment in first-person view and visualize robot-provided situational awareness overlays, as shown in Fig.  \ref{fig:flowchart}.

We define three coordinate frames:
\begin{itemize}
    \item $F_w$: World/map frame defined at robot initialization,
    \item $F_r$: Robot body frame,
    \item $F_{fpv}$: First-person LiDAR frame on the responder.
\end{itemize}

Rigid transforms are denoted ${}^aT_b(t) \in SE(3)$, mapping points from frame $F_b$ to frame $F_a$ at time $t$.  
At each timestamp $t$, the robot provides LiDAR packets $P^r_t$ and pose estimates ${}^wT_r(t)$, while the responder’s LiDAR produces local scans $P^{fpv}_t$.  
The task is threefold:

   \textbf{Drift-bounded robot odometry:}  
    Estimate ${}^wT_r(t)$ from robot LiDAR--IMU fusion to maintain a consistent global map $\mathcal{M}_w$. Prior to cross-LiDAR alignment, the global map $\mathcal{M}_w$ is pre-processed to remove ground and ceiling artifacts. 
Subsample points below a ground-near band and above a ceiling threshold are filtered, yielding a refined map $\mathcal{M}_w^{\ast}$ for robust registration.

    \textbf{Cross-LiDAR alignment:}  
    Compute the initial relative pose ${}^wT_{fpv}(t_0)$ between the responder LiDAR and the robot map $\mathcal{M}_w^{\ast}$ via scan registration (multi-resolution NDT followed by ICP).  
    The responder trajectory is then propagated as
    \begin{equation}
        {}^wT_{fpv}(t) = {}^wT_{fpv}(t_0) \cdot \Delta T_{fpv}(t_0 \!\to\! t),
    \end{equation}
    where $\Delta T_{fpv}$ is the relative motion from the responder’s onboard LIO.

   \textbf{AR projection:}  
Humans are detected from robot LiDAR scans as oriented 3D bounding boxes
\begin{equation}
\mathcal{B}_t = \{ B_k = (c_k, R_k, s_k) \mid c_k \in \mathbb{R}^3,\; R_k \in SO(3),\; s_k \in \mathbb{R}^3 \}.
\end{equation}
Each box $B_k$ is lifted to the world frame and projected into the responder’s view as
\begin{equation}
B^{fpv}_k = ({}^{fpv}T_w(t)) \, B_k.
\end{equation}

In addition, the subset of robot points belonging to detected humans is defined as
\begin{equation}
\mathcal{P}^r_{\text{human}}(t) = \{ p \in P^r_t \mid p \in B_k \;\; \text{for some } B_k \},
\end{equation}
and is projected into the responder’s view by
\begin{equation}
p^{fpv} = ({}^{fpv}T_w(t)) \, {}^wT_r(t) \, p, \quad p \in \mathcal{P}^r_{\text{human}}(t).
\end{equation}

Both $B^{fpv}_k$ and $\{p^{fpv}\}$ are rendered in red in the first-person AR display, enabling occlusion-aware situational overlays.

\subsection{System Initialization}

At deployment, both the mobile robot and the first-person (FPV) device run independent LiDAR--Inertial Odometry (LIO) pipelines. 
The initialization stage must establish (i) a global temporal reference across platforms, and (ii) a consistent spatial reference that anchors the FPV trajectory to the robot’s world frame.

\paragraph{Global Time Reference.}
We define $t=0$ as the instant when the robot LIO is launched, anchoring the world frame $F_w$ with ${}^wT_r(0) = I$. 
All computers are synchronized via PTP/NTP to bound inter-device clock skew, while each LiDAR packet is de-skewed using its own internal firing timestamps from the sensor driver. 
At a common render time $t_r$, both robot and FPV poses are interpolated to provide temporally consistent states, namely the robot world pose ${}^wT_r(t_r)$ and the FPV world pose ${}^wT_{fpv}(t_r)$.

\paragraph{World Frame Definition.}
Regardless of boot order, the global world frame $F_w$ is always anchored by the robot at its LIO initialization. 
If the FPV device starts earlier, its trajectory is first expressed in its own local frame with ${}^{fpv}T_{fpv}(t_0^{fpv}) = I$. 
Once the robot begins mapping, an inter-LiDAR registration establishes ${}^wT_{fpv}(t_0^{fpv})$, thereby anchoring the FPV trajectory to the robot-defined world frame. 
If the robot boots first, the alignment step simply associates the FPV odometry with the pre-defined $F_w$. 
Thus, the robot consistently serves as the global spatial reference for all subsequent mapping, alignment, and AR projection.

\paragraph{Dual LIO Boot.}
The robot runs FAST-LIO2 in ROS~2, publishing its trajectory ${}^wT_r(t)$ and incrementally building a voxelized NDT map $\mathcal{M}_w$. 
The FPV device runs its own LIO instance, outputting relative odometry increments $\Delta T_{fpv}(t_0^{fpv}\!\to\!t)$ referenced to its local origin. 
These two LIO streams evolve independently until inter-LiDAR registration is performed.

\paragraph{Cross-LiDAR Alignment.}
When sufficient robot map coverage is available in the region of interest, the FPV scan aggregated over $[t_0^{fpv}-\Delta,\, t_0^{fpv}]$ is downsampled and registered against $\mathcal{M}_w$. 
A multi-resolution NDT provides coarse initialization, followed by trimmed point-to-plane ICP under a Huber loss:
\begin{equation}
{}^wT_{fpv}(t_0^{fpv}) = 
\arg\min_{T \in SE(3)} \sum_{i=1}^N \rho\!\left(n_i^\top (T p_i - q_i)\right),
\end{equation}
where $p_i$ are FPV points, $q_i$ are map points, and $n_i$ their normals. 
The alignment is accepted only if residual and consistency metrics satisfy
\begin{equation}
\epsilon \leq \epsilon_{\max}, \quad 
\eta \geq \eta_{\min}, \quad 
\delta \leq \delta_{\max},
\end{equation}
where $\epsilon$ is RMS residual, $\eta$ the inlier ratio, and $\delta$ the update norm. 
If these conditions are not met, a yaw-grid reseed is attempted; otherwise initialization fails and FPV rendering is disabled until retry.

\paragraph{Pose Composition.}
Once accepted, the FPV world pose is maintained by composition:
\begin{equation}
{}^wT_{fpv}(t) = {}^wT_{fpv}(t_0^{fpv}) \cdot \Delta T_{fpv}(t_0^{fpv}\!\to\!t).
\end{equation}
This formulation accommodates asynchronous boot: regardless of when the FPV device is launched, its trajectory is aligned to the robot-defined world frame. 
All subsequent AR projection and rendering use this composed pose.

\subsection{Robot State Estimation}

The mobile robot defines and maintains the global world frame $F_w$. 
Its odometry is estimated by FAST-LIO2, a tightly coupled LiDAR--IMU method implemented in ROS~2. 
The robot state evolves through IMU-driven propagation
\begin{equation}
x_{t+1} = f(x_t, u_t) + w_t,
\end{equation}
and LiDAR updates
\begin{equation}
z_t = h(x_t, P^r_t) + v_t,
\end{equation}
where $u_t$ are inertial inputs and $P^r_t$ are LiDAR scans. 
The result is a drift-bounded trajectory ${}^wT_r(t)$ and a voxelized global map $\mathcal{M}_w$.

It is important to note that this global reference is unique to the robot: the FPV device runs its own 
independent LIO, which only estimates relative motion in its local coordinate frame. 
Therefore, the FPV trajectory cannot be directly expressed in $F_w$. 
To anchor it to the robot-defined world frame, an additional inter-LiDAR transformation 
${}^wT_{fpv}(t_0^{fpv})$ must be established via scan registration. 
This transform subsequently enables composition of the FPV trajectory in world coordinates, 
serving as the basis for consistent cross-view alignment and AR projection.

\paragraph{Pose Correction.}
While the initial registration ${}^wT_{fpv}(t_0^{fpv})$ anchors the FPV trajectory to the 
robot-defined world frame, accumulated drift in the FPV LIO may gradually degrade 
consistency. 
We assume that the majority of structural elements in the environment 
(e.g., walls, doors, furniture) remain static, and that no moving human subjects 
are present in the immediate field of view during correction. 
Under this assumption, the FPV pose can be periodically corrected by 
re-registering its LiDAR scans to the robot map.

Drift detection is based on registration quality indicators including the 
RMS residual $\epsilon$, inlier ratio $\eta$, and degeneracy index $\delta$. 
A pose correction is triggered whenever
\begin{equation}
\epsilon > \epsilon_{\max} \;\;\lor\;\; \eta < \eta_{\min} \;\;\lor\;\; \delta > \delta_{\max}.
\end{equation}
Once triggered, the FPV scan $P^{fpv}_t$ is aligned to the robot map $\mathcal{M}_w$ 
using the same NDT + ICP procedure as initialization:
\begin{equation}
{}^wT_{fpv}(t) \gets \arg\min_{T \in SE(3)} 
\sum_{i=1}^N \rho\!\left(n_i^\top (T p_i - q_i)\right).
\end{equation}
This pose correction mechanism ensures that, even under long-term operation 
and accumulated odometric drift, the FPV trajectory remains consistently 
anchored to the robot-defined world frame for reliable AR projection.

\subsection{Human Point Cloud Detection and AR Projection}

Based on the robot’s globally consistent trajectory and map, 
human observations can be detected and transferred into the FPV perspective.

\paragraph{Detection and Extraction.}
From the pointcloud $\mathcal{M}_w^{\ast}$, we adopt a PointPillars backbone 
implemented in \emph{OpenPCDet} to produce candidate oriented boxes 
$\mathcal{B}^r_t=\{B_k^r\}$ for the person class. 
These boxes are not directly rendered but serve as spatial filters to isolate 
the subset of points corresponding to human targets:
\begin{equation}
\mathcal{P}^r_{\text{human}}(t) = \{ p \in P^r_t \mid p \in B_k^r \;\; \text{for some } B_k^r \}.
\end{equation}

\paragraph{Projection to FPV.}
Each human point $p \in \mathcal{P}^r_{\text{human}}(t)$ is lifted to the world frame using the robot pose 
${}^wT_r(t)$ and then reprojected into the FPV frame:
\begin{equation}
p^{fpv} = ({}^{fpv}T_w(t)) \, {}^wT_r(t)\, p.
\end{equation}
The resulting set $\mathcal{P}^{fpv}_{\text{human}}(t)$ provides a dense first-person view of detected humans.

\paragraph{Rendering.}
For visualization, points in $\mathcal{P}^{fpv}_{\text{human}}(t)$ are colorized in red, while 
all other FPV points remain in default style:
\begin{equation}
\text{color}(p^{fpv}) =
\begin{cases}
\text{red}, & p^{fpv} \in \mathcal{P}^{fpv}_{\text{human}}(t), \\
\text{default}, & \text{otherwise}.
\end{cases}
\end{equation}
Bounding box wireframes or skeletal cues can optionally be projected as auxiliary hints, 
yet the colored point cloud naturally conveys the spatial extent of human presence in the FPV view.

\section{Experiments}

\noindent\textbf{Hardware Setup.}
Our experimental platform integrates both a first-person (FPV) device and a mobile robot, 
all implemented under ROS~2 framework.

\textit{FPV device.} 
A replica firearm is instrumented with a Picatinny-mounted Livox Avia LiDAR and 
a compact display for AR overlays. Processing and communication are handled by an 
NVIDIA Jetson Orin module carried in a backpack together with WiFi networking and 
a battery pack, enabling untethered operation for the first responder.

\textit{Mobile robot.} 
The ground unit is equipped with a Jetson Orin NX carrier board for onboard perception 
and runs FAST-LIO2 in real time. The robot chassis adopts an omnidirectional wheel 
design to allow agile maneuvering in confined spaces. The robot streams LiDAR, odometry, 
and map updates through ROS~2 nodes for cross-platform alignment and projection.

No additional sensors or hardware are required; all modules operate natively within 
the ROS~2 architecture.

\noindent \textbf{Ethics and Safety Statement.} 
No humans were harmed in this study. All operator-carried devices were safe props without weapon or water ejecting functionality. Human targets were represented by mannequins or volunteers under non-hazardous conditions. The goal of this work is solely to evaluate whether robots can assist in locating and projecting occluded humans into an AR view for 
safety-critical reconnaissance.

\begin{figure}[h]
  \centering
  \includegraphics[width=\linewidth]{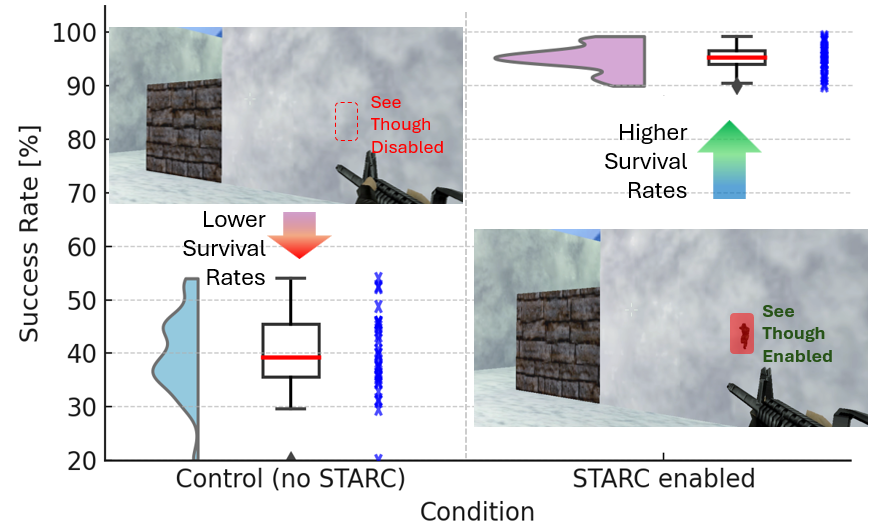}
  \vspace{-5pt}
  \caption{Survival rate comparison between baseline and STARC conditions. Left: without STARC, participants must operate under occlusion, leading to blind spots (see-through disabled) and lower survival rates. Right: with STARC enabled, detected humans are projected through occluding structures (see-through enabled), yielding higher survival rates. Half-violin plots show the underlying distribution, boxplots highlight median and quartiles, and blue dots represent individual participant trials.}
  \label{fig:simulation_result}
\end{figure}
\subsection{Experimental Objectives.}
There is \textbf{no directly comparable prior system}, as STARC is the \textbf{first framework} to combine \textbf{dual LIO}, \textbf{cross-LiDAR alignment}, and \textbf{first-person AR overlays}; hence, our evaluation focuses on validating both \textbf{task-level} and \textbf{system-level} performance. The experiments are designed to validate STARC from both a \textbf{task-level} and \textbf{system-level} perspective. At the \textbf{task level}, we measure whether \textbf{see-through AR overlays} improve operator performance in adversarial scenarios, quantified by \textbf{engagement outcomes (successes vs. failures)} under a controlled simulation environment. At the \textbf{system level}, we quantify the \textbf{accuracy} and \textbf{efficiency} of \textbf{through-door projection} in lab settings, using \textbf{inlier/outlier ratios} against RGB-SAM masks and \textbf{end-to-end latency} (including SLAM, communication, and rendering). Together, these evaluations provide evidence that STARC not only enhances \textbf{mission effectiveness} for human operators, but also maintains the \textbf{accuracy} and \textbf{runtime feasibility} required for \textbf{real deployment}.

\begin{figure*}[t]
  \centering
  \includegraphics[width=\linewidth]{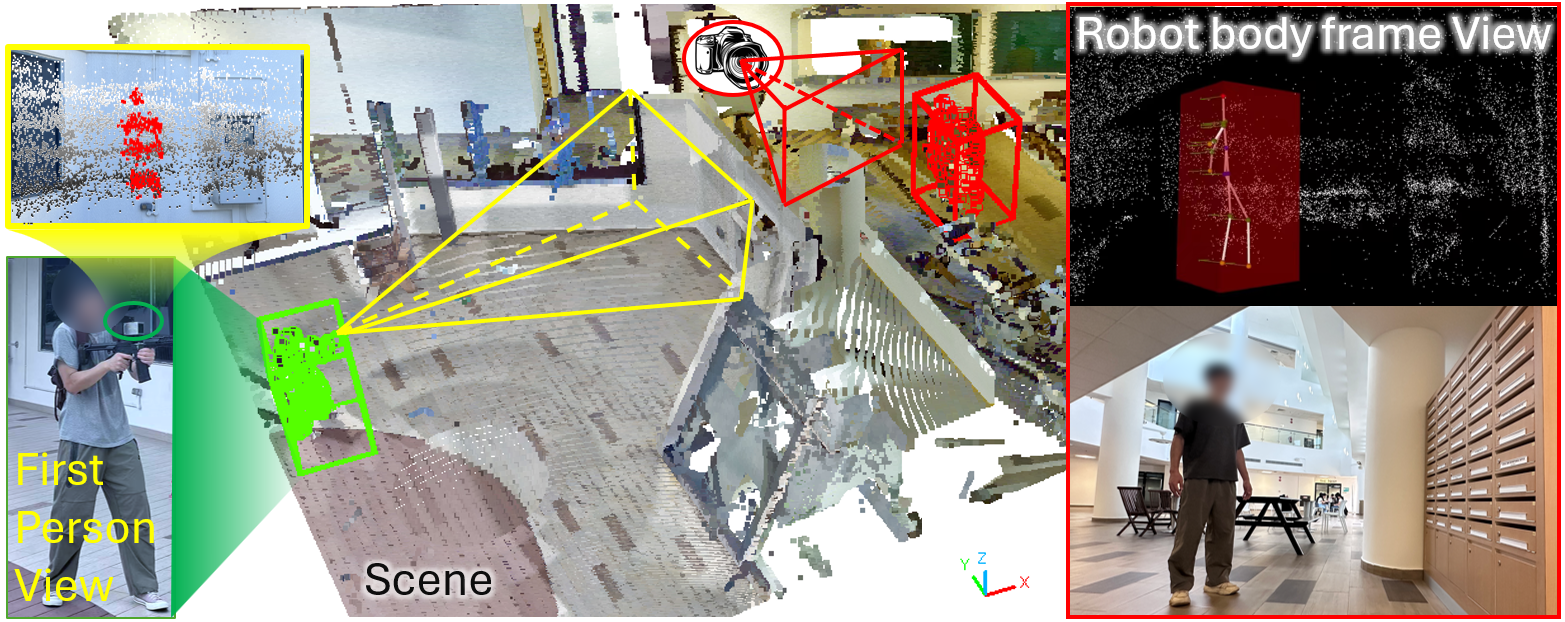}
  \vspace{-8pt}
  \caption{Human–robot cooperative perception in an emergency response scenario. The robot constructs a global LiDAR map and detects humans (red boxes), while the responder’s first-person LiDAR is aligned for real-time projection of detections into the AR view, enabling see-through situational awareness under occlusion.}
  \label{fig:result_ntutitive}
    \vspace{-8pt}
\end{figure*}

\subsection{Simulation Evaluation.}
To complement the real-world experiments, we conducted controlled trials in a simulated adversarial environment. A modified first-person shooter engine was used to emulate one-versus-one engagements against a high-difficulty opponent. Two vision conditions were compared: (i) baseline vision without AR overlays, and (ii) STARC-enabled vision with see-through projection. To isolate the effect of AR support from system latency, the simulation assumed be zero delay in perception and rendering.

Thirty participants were recruited: one police cadet collaborating with the local research department, ten self-identified pro gamers, and nineteen casual users. Each participant performed at least ten rounds per condition. Some of the initial participants tested for 100 rounds. But the task is simple and quickly learned. More trials would primarily lead to overfitting to the game mechanics rather than revealing meaningful differences. That's why in the later tests, it was reduced to at least 10 rounds. 

For every round, outcomes were logged as either a success (opponent eliminated) or a failure (participant eliminated). As shown in Fig. \ref{fig:simulation_result}, with STARC enabled, nearly all gamer participants achieved consistently high survival rates, demonstrating the effectiveness of see-through AR support. The police cadet, however, showed relatively lower performance: both a lack of gaming experience and professional training that discourages firing through walls reduced their ability to exploit the AR overlays. This underscores that while STARC reliably improves situational awareness, ultimate task outcomes remain strongly influenced by the operator’s baseline proficiency in the environment.

\subsection{Real World Evaluation in Controlled Settings}
\begin{figure*}[t]
  \centering
  \includegraphics[width=\linewidth]{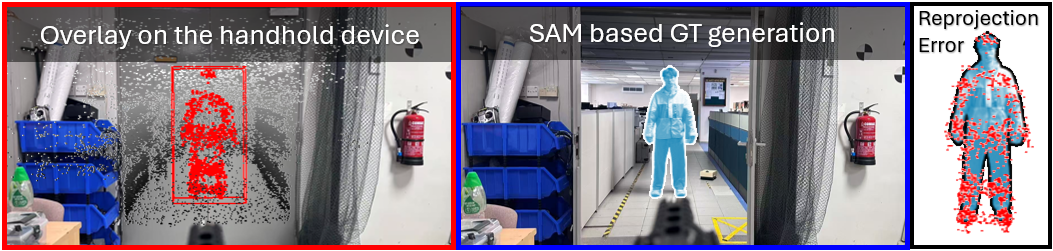}
    \vspace{-12pt}
 \caption{Quantitative evaluation in controlled indoor settings. (Left) All projected LiDAR points (red) overlaid on the responder’s handheld device view. (Middle) Ground-truth human mask obtained by SAM segmentation from synchronized RGB frames. (Right) Reprojection error visualization: points outside the mask contour are counted as outliers (red), while points inside are considered inliers.}
  \label{fig:Evaluation_result}
    \vspace{-12pt}
\end{figure*}

In controlled experiments (See Fig.  \ref{fig:Evaluation_result} and Tab. \ref{tab:sam_eval}), we evaluate through-door projection by comparing projected 
human point clouds against 2D segmentation masks obtained from RGB frames. 
After opening the door, SAM is applied to the RGB stream to produce human masks 
$\mathcal{M}_{SAM}$. Projected LiDAR points 
$\mathcal{P}^{fpv}_{\text{human}}=\{p_i\}_{i=1}^{N}$ are re-projected into the RGB 
image plane via extrinsic calibration. The inlier ratio is then defined as
\begin{equation}
r_{\text{inlier}} = 
\frac{1}{N}\sum_{i=1}^N \mathbf{1}\!\left[p_i \in \mathcal{M}_{SAM}\right], r_{\text{outlier}} = 1 - r_{\text{inlier}},
\end{equation}
where $\mathbf{1}[\cdot]$ is the indicator function.
In addition, we report the end-to-end latency $L_{\text{tot}}$, which accounts for 
robot SLAM ($L_{\text{SLAM}}$), inter-device communication ($L_{\text{comm}}$), 
and AR rendering ($L_{\text{viz}}$):
\begin{equation}
L_{\text{tot}} = L_{\text{SLAM}} + L_{\text{comm}} + L_{\text{viz}}.
\end{equation}
These metrics jointly quantify both overlay accuracy and runtime feasibility.

\begin{table}[h]
\centering
\caption{Evaluation of projected human point clouds against RGB-SAM masks. 
Metrics include inlier and outlier ratios and end-to-end latency. }
\label{tab:sam_eval}
\begin{tabular}{lccc}
\hline
\toprule
Scenario & Inlier [\%] $\uparrow$ & Outlier [\%] $\downarrow$ & $L_{\text{tot}}$ [ms] $\downarrow$ \\
\midrule
Auditorium      & 87.1 & 12.9 & 111.6 \\
Lab Scene       & 89.5 & 10.5 & 42.0 \\
Tactical site   & 82.4 & 17.6 & 57.7 \\
\midrule
Mean $\pm$ Std  & 86.3 $\pm$ 2.9 & 13.7 $\pm$ 2.9 & 70.4 $\pm$ 29.8 \\
\bottomrule
\end{tabular}
\caption*{\footnotesize Note: The total time $L_{\text{tot}}$ is counted from input to the LIO through to projection at the display.}
\end{table}

\noindent\textit{Discussion.}
Across all scenarios, more than 80\% of projected human points consistently fall inside 
the RGB-SAM masks, with outliers bounded below 20\%. 
The total end-to-end latency in ideal case remains within $\sim$50\,ms, including SLAM, 
communication, and AR rendering, confirming that the through-door overlays 
are both spatially consistent and real-time deployable.

\subsection{Real-World Field Trials}
Beyond simulation and controlled lab settings, we validate STARC in two representative 
real-world environments. 

\textit{Auditorium.} 
The first deployment takes place in an auditorium with multiple entry doors 
and occluded seating areas (see Fig.  \ref{fig:result_ntutitive}). The robot performs an initial frontier-based exploration to 
construct a LiDAR map of the interior, while the FPV device remains outside. When the 
operator enters, through-wall projections of occluded humans are rendered in the 
first-person AR view, enabling awareness of hidden occupants before line-of-sight is 
established.

\textit{Tactical training center.} 
We further evaluate STARC in a dedicated tactical facility designed for first-responder 
exercises. The environment includes partitioned rooms, corridors, and staged targets. 
The robot is remotely driven to scan the interior, and FPV overlays provide the operator 
with situational awareness of terrorists and hostages concealed behind walls. 
Fig. ~\ref{fig:motivation} illustrates the tactical setup, while 
Fig. ~\ref{fig:simulation_result} shows the resulting overlays in both world-centric 
and first-person views.

These trials demonstrate that STARC is not limited to simplified lab scenes but can 
operate under realistic indoor layouts, confirming its readiness for safety-critical 
reconnaissance tasks.

\section{Limitations}

\noindent\textbf{System dependencies.}
Our prototype hinges on three coupled modules. First, inter-LIO registration anchors FPV trajectories to the robot map; without it, cross-LiDAR alignment and AR projection fail. Second, stable odometry from FAST-LIO2 is essential, as feature-scarce scenes (e.g., planar halls) can cause drift that disrupts mapping and human detection. Third, low-latency links are required; degraded communication prevents projection and undermines awareness. These dependencies expose current limitations but also suggest extensions such as redundant anchors, multi-perspective sensing, and comm-aware planning.

\noindent\textbf{Parameter tuning.}
System parameters (e.g., ICP thresholds, filter cutoffs, re-alignment triggers) are tuned empirically to ensure robust performance in our evaluated environments. 
While effective for proof-of-concept validation, broader deployment may benefit from adaptive or learned policies. 
Exploring reinforcement learning or other data-driven approaches for online parameter optimization is a promising avenue for future work.

\section{Conclusion}

We presented a first-person see-through AR pipeline for safety-critical indoor operations built on a \emph{dual LIO} design: a robot runs FAST-LIO2 to maintain a voxelized world map, while a handhold/helmet-mounted FPV device runs its own LIO, with one-time inter-LIO alignment enabling consistent projection of 3D human detections (PointPillars) into the operator’s AR view under strict latency budgets. Implemented in ROS~2 with explicit time synchronization, packet de-skew, and deterministic re-alignment policies, the system was validated across a tactical training site, a large auditorium, and a lab corridor with ground truth, showing stable overlays, accurate FPV poses, and superior robustness over scan-to-map baselines. As a proof-of-concept, STARC does not aim to introduce new perception algorithms but to demonstrate the first ROS~2-native, real-time integration of cross-LiDAR alignment and AR overlays for occlusion-aware situational awareness, providing a critical step toward deployable systems that can later incorporate loop closures, heterogeneous robot–device teams, and higher-level human activity recognition. Our code and design will be open source upon acceptance.

\section*{Acknowledgments} 
The authors acknowledge the use of large language models (i.e., ChatGPT by OpenAI) for assisting in prototype code generation and improving the clarity of the manuscript. All technical contributions and interpretations remain the responsibility of the authors.

\bibliographystyle{IEEEtran}
\bibliography{mybib}

\begin{thebibliography}{10}
\providecommand{\url}[1]{#1}
\csname url@samestyle\endcsname
\providecommand{\newblock}{\relax}
\providecommand{\bibinfo}[2]{#2}
\providecommand{\BIBentrySTDinterwordspacing}{\spaceskip=0pt\relax}
\providecommand{\BIBentryALTinterwordstretchfactor}{4}
\providecommand{\BIBentryALTinterwordspacing}{\spaceskip=\fontdimen2\font plus
\BIBentryALTinterwordstretchfactor\fontdimen3\font minus \fontdimen4\font\relax}
\providecommand{\BIBforeignlanguage}[2]{{%
\expandafter\ifx\csname l@#1\endcsname\relax
\typeout{** WARNING: IEEEtran.bst: No hyphenation pattern has been}%
\typeout{** loaded for the language `#1'. Using the pattern for}%
\typeout{** the default language instead.}%
\else
\language=\csname l@#1\endcsname
\fi
#2}}
\providecommand{\BIBdecl}{\relax}
\BIBdecl

\bibitem{schroth2024emergency}
C.~A. Schroth, C.~Eckrich, I.~Kakouche, S.~Fabian, O.~Von~Stryk, A.~M. Zoubir, and M.~Muma, ``Emergency response person localization and vital sign estimation using a semi-autonomous robot mounted sfcw radar,'' \emph{IEEE Transactions on Biomedical Engineering}, vol.~71, no.~6, pp. 1756--1769, 2024.

\bibitem{erat2018drone}
O.~Erat, W.~A. Isop, D.~Kalkofen, and D.~Schmalstieg, ``Drone-augmented human vision: Exocentric control for drones exploring hidden areas,'' \emph{IEEE transactions on visualization and computer graphics}, vol.~24, no.~4, pp. 1437--1446, 2018.

\bibitem{karanam20173d}
C.~R. Karanam and Y.~Mostofi, ``3d through-wall imaging with unmanned aerial vehicles using wifi,'' in \emph{Proceedings of the 16th ACM/IEEE International Conference on Information Processing in Sensor Networks}, 2017, pp. 131--142.

\bibitem{chen20243d}
J.~Chen, B.~Sun, M.~Pollefeys, and H.~Blum, ``A 3d mixed reality interface for human-robot teaming,'' in \emph{2024 IEEE International Conference on Robotics and Automation (ICRA)}.\hskip 1em plus 0.5em minus 0.4em\relax IEEE, 2024, pp. 11\,327--11\,333.

\bibitem{sun2025frontiernet}
B.~Sun, H.~Chen, S.~Leutenegger, C.~Cadena, M.~Pollefeys, and H.~Blum, ``Frontiernet: Learning visual cues to explore,'' \emph{IEEE Robotics and Automation Letters}, 2025.

\bibitem{yue2022aerial}
Y.~Yue, C.~Zhao, Y.~Wang, Y.~Yang, and D.~Wang, ``Aerial-ground robots collaborative 3d mapping in gnss-denied environments,'' in \emph{2022 International Conference on Robotics and Automation (ICRA)}.\hskip 1em plus 0.5em minus 0.4em\relax IEEE, 2022, pp. 10\,041--10\,047.

\bibitem{li2024colag}
Z.~Li, R.~Mao, N.~Chen, C.~Xu, F.~Gao, and Y.~Cao, ``Colag: A collaborative air-ground framework for perception-limited ugvs’ navigation,'' in \emph{2024 IEEE International Conference on Robotics and Automation (ICRA)}.\hskip 1em plus 0.5em minus 0.4em\relax IEEE, 2024, pp. 16\,781--16\,787.

\bibitem{coffey2022collaborative}
M.~Coffey and A.~Pierson, ``Collaborative teleoperation with haptic feedback for collision-free navigation of ground robots,'' in \emph{2022 IEEE/RSJ International Conference on Intelligent Robots and Systems (IROS)}.\hskip 1em plus 0.5em minus 0.4em\relax IEEE, 2022.

\bibitem{qiu2020shared}
S.~Qiu, H.~Liu, Z.~Zhang, Y.~Zhu, and S.-C. Zhu, ``Human–robot interaction in a shared augmented reality workspace,'' in \emph{2020 IEEE/RSJ International Conference on Intelligent Robots and Systems (IROS)}.\hskip 1em plus 0.5em minus 0.4em\relax IEEE, 2020, pp. 11\,413--11\,418.

\bibitem{xu2022fast}
W.~Xu, Y.~Cai, D.~He, J.~Lin, and F.~Zhang, ``Fast-lio2: Fast direct lidar-inertial odometry,'' \emph{IEEE Transactions on Robotics}, vol.~38, no.~4, pp. 2053--2073, 2022.

\bibitem{chen2024ig}
Z.~Chen, Y.~Xu, S.~Yuan, and L.~Xie, ``ig-lio: An incremental gicp-based tightly-coupled lidar-inertial odometry,'' \emph{IEEE Robotics and Automation Letters}, vol.~9, no.~2, pp. 1883--1890, 2024.

\bibitem{li2024hcto}
J.~Li, S.~Yuan, M.~Cao, T.-M. Nguyen, K.~Cao, and L.~Xie, ``Hcto: Optimality-aware lidar inertial odometry with hybrid continuous time optimization for compact wearable mapping system,'' \emph{ISPRS Journal of Photogrammetry and Remote Sensing}, vol. 211, pp. 228--243, 2024.

\bibitem{li2025ua}
J.~Li, X.~Xu, J.~Liu, K.~Cao, S.~Yuan, and L.~Xie, ``Ua-mpc: Uncertainty-aware model predictive control for motorized lidar odometry,'' \emph{IEEE Robotics and Automation Letters}, 2025.

\bibitem{li2025graph}
J.~Li, T.-M. Nguyen, M.~Cao, S.~Yuan, T.-Y. Hung, and L.~Xie, ``Graph optimality-aware stochastic lidar bundle adjustment with progressive spatial smoothing,'' \emph{IEEE Transactions on Intelligent Transportation Systems}, 2025.

\bibitem{zhu2023swarmlio}
F.~Zhu, Y.~Ren, F.~Kong, H.~Wu, S.~Liang, N.~Chen, W.~Xu, and F.~Zhang, ``Swarm-lio: Decentralized swarm lidar-inertial odometry,'' in \emph{2023 IEEE International Conference on Robotics and Automation (ICRA)}, 2023, pp. 3254--3260.

\bibitem{zhu2023swarmlio2}
F.~Zhu, Y.~Ren, L.~Yin, F.~Kong, Q.~Liu, R.~Xue, W.~Liu, Y.~Cai, G.~Lu, H.~Li, and F.~Zhang, ``Swarm-lio2: Decentralized efficient lidar-inertial odometry for aerial swarm systems,'' \emph{IEEE Transactions on Robotics}, vol.~41, pp. 960--981, 2025.

\bibitem{biber2003normal}
P.~Biber and W.~Stra{\ss}er, ``The normal distributions transform: A new approach to laser scan matching,'' in \emph{Proceedings 2003 IEEE/RSJ International Conference on Intelligent Robots and Systems (IROS 2003)(Cat. No. 03CH37453)}, vol.~3.\hskip 1em plus 0.5em minus 0.4em\relax IEEE, 2003, pp. 2743--2748.

\bibitem{magnusson2009evaluation}
M.~Magnusson, A.~Nuchter, C.~Lorken, A.~J. Lilienthal, and J.~Hertzberg, ``Evaluation of 3d registration reliability and speed-a comparison of icp and ndt,'' in \emph{2009 IEEE International Conference on Robotics and Automation}.\hskip 1em plus 0.5em minus 0.4em\relax IEEE, 2009, pp. 3907--3912.

\bibitem{yuan2023STD}
C.~Yuan, J.~Lin, Z.~Zou, X.~Hong, and F.~Zhang, ``Std: Stable triangle descriptor for 3d place recognition,'' in \emph{2023 IEEE International Conference on Robotics and Automation (ICRA)}, 2023, pp. 1897--1903.

\bibitem{yin2024outram}
P.~Yin, H.~Cao, T.-M. Nguyen, S.~Yuan, S.~Zhang, K.~Liu, and L.~Xie, ``Outram: One-shot global localization via triangulated scene graph and global outlier pruning,'' in \emph{2024 IEEE International Conference on Robotics and Automation (ICRA)}.\hskip 1em plus 0.5em minus 0.4em\relax IEEE, 2024, pp. 13\,717--13\,723.

\bibitem{Yin2023segregator}
P.~Yin, S.~Yuan, H.~Cao, X.~Ji, S.~Zhang, and L.~Xie, ``Segregator: Global point cloud registration with semantic and geometric cues,'' in \emph{2023 IEEE International Conference on Robotics and Automation (ICRA)}, 2023, pp. 2848--2854.

\bibitem{kim2018scan}
G.~Kim and A.~Kim, ``Scan context: Egocentric spatial descriptor for place recognition within 3d point cloud map,'' in \emph{2018 IEEE/RSJ International Conference on Intelligent Robots and Systems (IROS)}.\hskip 1em plus 0.5em minus 0.4em\relax IEEE, 2018, pp. 4802--4809.

\bibitem{luo2023bevplace}
L.~Luo, S.~Zheng, Y.~Li, Y.~Fan, B.~Yu, S.-Y. Cao, J.~Li, and H.-L. Shen, ``Bevplace: Learning lidar-based place recognition using bird's eye view images,'' in \emph{Proceedings of the IEEE/CVF International Conference on Computer Vision}, 2023, pp. 8700--8709.

\bibitem{cai2025bevlio}
H.~Cai, S.~Yuan, X.~Li, J.~Guo, and J.~Liu, ``Bev-lio (lc): Bev image assisted lidar-inertial odometry with loop closure,'' in \emph{Proceedings of the IEEE/RSJ International Conference on Intelligent Robots and Systems (IROS)}, 2025, to appear.

\bibitem{Shi_2020_CVPR}
S.~Shi, C.~Guo, L.~Jiang, Z.~Wang, J.~Shi, X.~Wang, and H.~Li, ``Pv-rcnn: Point-voxel feature set abstraction for 3d object detection,'' in \emph{Proceedings of the IEEE/CVF conference on computer vision and pattern recognition}, 2020, pp. 10\,529--10\,538.

\bibitem{Yin_2021_CVPR}
T.~Yin, X.~Zhou, and P.~Krahenbuhl, ``Center-based 3d object detection and tracking,'' in \emph{Proceedings of the IEEE/CVF conference on computer vision and pattern recognition}, 2021, pp. 11\,784--11\,793.

\bibitem{Lang_2019_CVPR}
A.~H. Lang, S.~Vora, H.~Caesar, L.~Zhou, J.~Yang, and O.~Beijbom, ``Pointpillars: Fast encoders for object detection from point clouds,'' in \emph{Proceedings of the IEEE/CVF conference on computer vision and pattern recognition}, 2019, pp. 12\,697--12\,705.

\bibitem{Qin_2022_ECCV}
Y.~Qin, C.~Wang, Z.~Kang, N.~Ma, Z.~Li, and R.~Zhang, ``Supfusion: Supervised lidar-camera fusion for 3d object detection,'' in \emph{Proceedings of the IEEE/CVF international conference on computer vision}, 2023, pp. 22\,014--22\,024.

\bibitem{Luo_2024_IROS}
R.~Luo, M.~Zolotas, D.~Moore, and T.~Pad{\i}r, ``User-customizable shared control for robot teleoperation via virtual reality,'' in \emph{2024 IEEE/RSJ International Conference on Intelligent Robots and Systems (IROS)}.\hskip 1em plus 0.5em minus 0.4em\relax IEEE, 2024, pp. 12\,196--12\,203.

\bibitem{walker2021mixed}
M.~Walker, Z.~Chen, M.~Whitlock, D.~Blair, D.~A. Szafir, C.~Heckman, and D.~Szafir, ``A mixed reality supervision and telepresence interface for outdoor field robotics,'' in \emph{2021 IEEE/RSJ International Conference on Intelligent Robots and Systems (IROS)}.\hskip 1em plus 0.5em minus 0.4em\relax IEEE, 2021, pp. 2345--2352.

\end{thebibliography}


\end{document}